# Think as Needed: Geometry-Driven Adaptive Perception for Autonomous Driving

*Enhanced HOPE: Adaptive Grassmannian Perception with Episodic Memory via Local Intrinsic Dimension Estimation*

**Donghyun Kim**
Stony Brook University
donghyun.kim.7@stonybrook.edu

**Jaehyoung Park**
Stony Brook University
jaehyoung.park@stonybrook.edu

## Abstract

Autonomous driving scenes range from empty highways to dense intersections with dozens of interacting road users, yet current 3D detection models apply a fixed computation budget to every frame—wasting resources on simple scenes while lacking capacity for complex ones. Existing approaches compound this problem: Transformer-based interaction models scale quadratically with the number of detected objects, and frame-by-frame processing causes the system to immediately forget objects the moment they become occluded. We propose **Enhanced HOPE**, an adaptive perception architecture that measures the geometric complexity of each incoming LiDAR frame using an unsupervised statistical estimator and routes it through a shallow or deep processing path accordingly, requiring no manual scene labels. To keep interaction modeling efficient, we replace quadratic pairwise attention with a linear-time subspace-based network that groups nearby objects into clusters and processes them jointly. The computational savings from these two mechanisms free up resources for a persistent temporal memory module that retains previously detected objects and traffic rules across frames, enabling the system to recall occluded objects seconds after they disappear from view. On the nuScenes and CARLA benchmarks, Enhanced HOPE reduces latency by 38% on simple scenes with no accuracy loss, improves mean Average Precision by 2.7 points on rare long-tail scenarios, and tracks objects through occlusions lasting over 5 seconds—where all tested baselines fail.

## 1 Introduction

A Level 4+ autonomous driving system must perceive its environment across an enormous range of traffic situations. This diversity in scene complexity and agent density is the central challenge: on an empty highway, only a handful of distant vehicles need to be tracked, whereas at a busy urban intersection, the perception system must simultaneously detect, classify, and predict the behavior of dozens of agents—the vehicles, pedestrians, cyclists, and other road users extracted from the LiDAR point cloud.

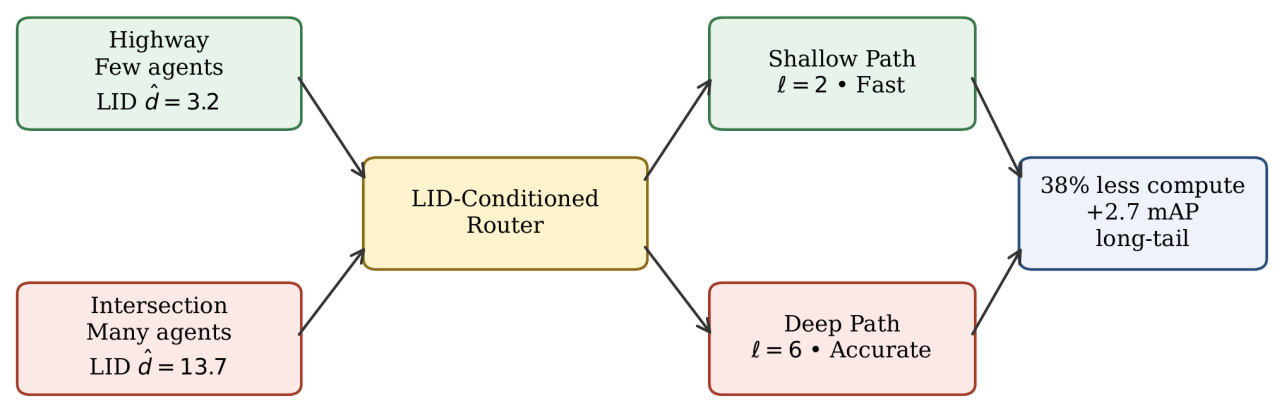

**Figure 1: Key idea.** Enhanced HOPE estimates the geometric complexity of each LiDAR frame using Local Intrinsic Dimension (LID) and routes it to a shallow (fast) or deep (accurate) processing path. Simple scenes save 38% compute; complex scenes gain +2.7 mAP.

Despite this diversity, current perception architectures apply the same neural network at full depth to every frame. This fixed-compute design creates three concrete problems:

**(1) Wasted computation.** On easy frames (e.g., highway driving), the full model runs unnecessary operations, wasting latency and energy—critical resources on embedded automotive hardware [13, 15].

**(2) Quadratic scaling.** Most state-of-the-art models use Transformer attention [23] to model interactions between agents. However, attention compares every agent to every other agent, scaling as $O(L^2)$ with the agent count $L$. In a crowded intersection with $L > 100$, this becomes a critical bottleneck.

**(3) Occlusion amnesia.** Standard models process each frame independently and "forget" objects the moment they become occluded. A construction sign hidden behind a truck for a few seconds is lost from the system's state, even though the ego vehicle will reach the construction zone moments later.

These three problems are not independent—they are deeply linked. We observe that *efficiency buys intelligence*: if the system can avoid wasting computation on simple scenes and can model agent interactions cheaply, the resulting computational headroom can be invested in expensive but high-value capabilities like persistent memory. This insight drives the design of Enhanced HOPE (Figure 1), an adaptive architecture in which each component directly addresses one of the three problems, and the savings from the first two enable the third.

**Problem 1 → LID-based Adaptive Routing.** We let the geometry of the data itself decide how much computation each frame deserves. Specifically, we estimate the Local Intrinsic

Dimension (LID) of each incoming point cloud—a well-studied statistical quantity that measures the effective number of independent degrees of freedom in the data [9]. A highway point cloud, with a few agents moving along a single axis, produces a low LID; a busy intersection, with agents moving in many directions, produces a high LID. This single number routes the frame through a shallow (fast) or deep (thorough) processing path, without any supervised complexity labels. Easy frames consume up to 38% fewer operations; hard frames receive the full model capacity.

**Problem 2 → Subspace-based Interaction Network.** Rather than comparing all $L$ agents pairwise (as in Transformer attention), we model their interactions using a Grassmannian Hypergraph Network (GHN). The intuition is twofold. First, each agent is represented not as a single feature vector but as a low-dimensional subspace—a compact summary of its multiple behavior modes (accelerating, braking, turning)—living on the Grassmann manifold [14], a mathematical space whose "points" are themselves subspaces of $\mathbb{R}^n$. Second, instead of connecting agents in pairs, we group spatially nearby agents into overlapping clusters called hyperedges [10] and process each cluster jointly, achieving $O(L)$ message passing instead of $O(L^2)$ attention. The result is a backbone that is both more expressive (capturing group dynamics) and cheaper to run.

**Problem 3 → Persistent Temporal Memory.** The computational budget freed by adaptive routing and linear-time interaction modeling is invested in a persistent temporal memory module, adapted from the Titans architecture [4] and the HOPE continual learning module from the Nested Learning framework [3]. The memory operates on two timescales: a short-term cross-attention buffer that tracks recent object states across frames, and a long-term neural memory that absorbs recurring traffic rules and road geometries. Together, they allow the system to recall occluded objects and temporary regulations (e.g., construction-zone speed limits) long after they leave the sensor's field of view. Without the savings from the first two components, this module would be too expensive to run at real-time rates.

**Contributions.** Our work makes the following contributions, organized around a central thesis: *geometry-aware adaptive computation enables efficient, robust, and memory-capable perception for autonomous driving.*

- **(Main) LID-conditioned adaptive routing** (Section 3.1): the first use of intrinsic dimension as an unsupervised, online complexity signal to dynamically adjust model depth in autonomous driving perception. This is the primary architectural innovation that enables the efficiency gains underlying the entire system.
- **(Efficiency backbone) Grassmannian Hypergraph Network** (Sections 3.2 and 3.3): a subspace-based interaction model with provably $O(L)$ complexity that replaces $O(L^2)$ Transformer attention, providing the linear-time backbone within which adaptive routing operates.
- **(Robustness module) Persistent temporal memory** (Section 3.4): a dual-timescale memory adapted from language modeling to the spatio-temporal domain, made feasible by the computational headroom from the above two components, enabling persistent object tracking through prolonged occlusions.
- **Empirical validation on nuScenes** [7] **and a custom CARLA-LongTail benchmark**: 38% latency reduction on simple scenes, +2.7 mAP on long-tail scenarios, and +22.6 points in occlusion tracking vs. the strongest baseline.

## 2 Related Work

**End-to-End Autonomous Driving.** UniAD [13] unifies detection, tracking, and planning but uses static computation. VAD [15] and SparseDrive [21] improve efficiency but lack complexity-aware routing. Enhanced HOPE's adaptive mechanism is complementary to all of these.

**Adaptive Computation.** Early-exit networks [22] and MoE [18] condition adaptation on learned gating, which is brittle under distribution shift. Recent State Space Models such as Mamba [11] achieve $O(L)$ complexity for long sequences via selective state spaces, offering an alternative to quadratic attention. However, SSMs provide a fixed linear-time backbone and do not adapt model depth or capacity to input complexity. Enhanced HOPE conditions adaptation on a geometric invariant (LID), dynamically routing inputs through variable-depth paths—an orthogonal and complementary strategy.

**Nested Learning & HOPE.** Behrouz et al. [3] introduced Nested Learning (NL), a paradigm representing ML models as nested, multi-level optimization problems with their own context flows. They showed that gradient-based optimizers are associative memory modules, and proposed a self-modifying sequence model combined with a continuum memory system, forming the HOPE module. HOPE demonstrated strong results in language modeling, continual learning, and long-context reasoning. We extend the HOPE philosophy—self-modification and continuum memory—to autonomous driving perception, replacing the language-oriented components with a Grassmannian hypergraph formulation and LID-conditioned routing for real-time sensor data.

**Titans Architecture.** Titans [4] introduced a neural long-term memory module that learns to memorize historical context at test time. From a memory perspective, attention serves as short-term memory (limited context, accurate dependencies), while the neural memory module acts as long-term, persistent memory. Titans demonstrated effective scaling beyond 2M context windows with superior performance on language modeling, commonsense reasoning, genomics, and time series tasks. We adapt the Titans memory architecture for spatio-temporal traffic context, introducing a Contextual Gating mechanism for sensor-noise robustness.

**Geometric Deep Learning & Hypergraph Networks.** Bronstein et al. [6] established a unifying framework for geometric deep learning across grids, groups, graphs, and geodesics. GrNet [14] applies Grassmannian geometry to video classification, and Grassmann pooling has been used in few-shot learning [19]. Generalizing standard GNNs, Hypergraph Neural Networks [10] extend message passing to hyperedges that capture higher-order correlations beyond pairwise interactions. We combine Grassmannian subspace modeling with hypergraph message passing à la Feng et al. [10] to efficiently capture multi-agent group dynamics in traffic scenes.

## 3 Method

Figure 2 provides an overview of Enhanced HOPE.

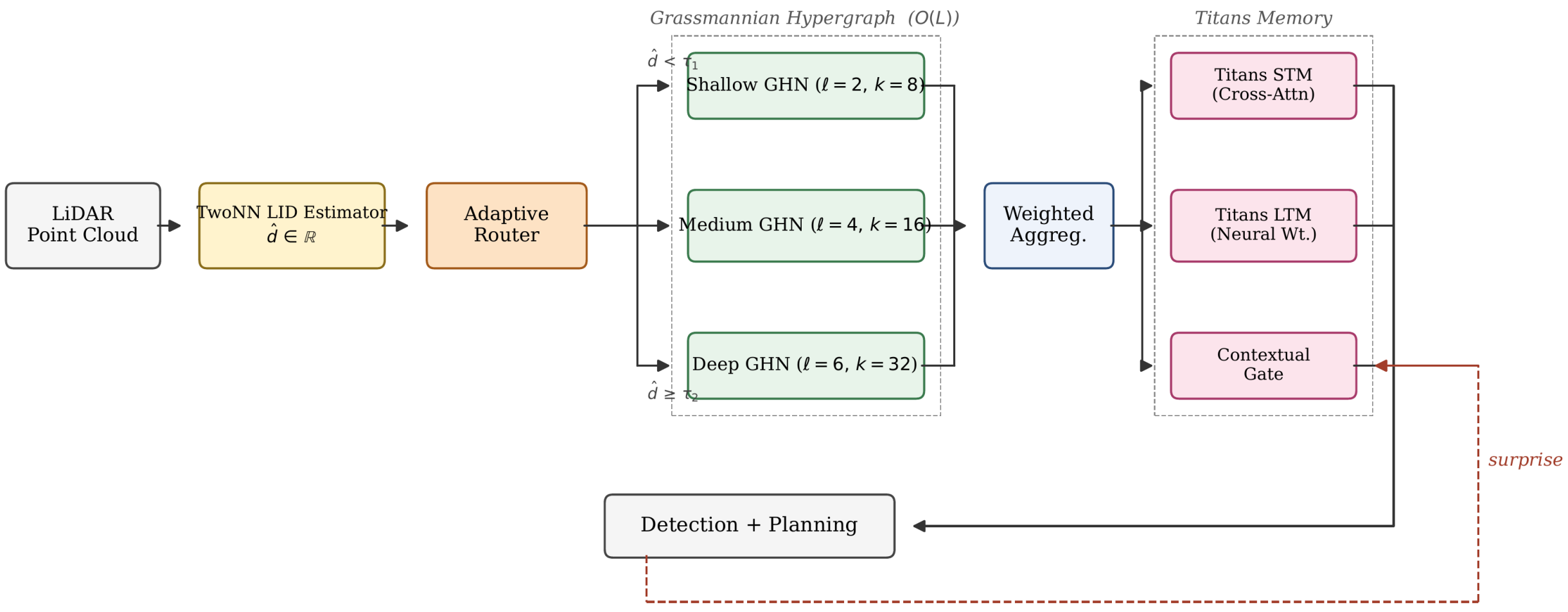

**Figure 2: Enhanced HOPE Architecture.** The TwoNN LID estimator routes the point cloud through an adaptive-depth Grassmannian Hypergraph Network (GHN). Features are then passed to a Titans-based episodic memory (STM+LTM) with contextual gating for persistent object tracking.

### 3.1 Local Intrinsic Dimension Monitor

Given a LiDAR point cloud with $L$ points, we need a fast, assumption-free estimate of its geometric complexity. We adopt the TwoNN estimator [9], which computes the LID $\hat{d}$ from the ratio of second-to-first nearest-neighbor distances across all points. Intuitively, $\hat{d}$ measures the effective number of independent degrees of freedom in the local geometry: a highway scene with cars moving along a single lane yields $\hat{d} \approx 3$, while a busy intersection with agents moving in many directions yields $\hat{d} \approx 14$. The estimator runs in amortized $O(L)$ time using a KD-tree over voxelized points.

We use $\hat{d}$ as the input to a differentiable soft routing gate that distributes computation across $J = 3$ processing paths (shallow, medium, deep):

$$w_j = exp(-\beta\,|\hat{d} - c_j|) \;/\; \textstyle\sum_{j'} exp(-\beta\,|\hat{d} - c_{j'}|) \tag{1}$$

where $c_j$ are learnable path centers and $\beta$ is a temperature parameter annealed during training. At inference, we use a hard $\arg\max$ with the straight-through estimator [5] for gradient propagation. This is our core mechanism: the LID signal decides how much the network should think about each frame, without any supervised complexity annotations.

### 3.2 Grassmannian Hypergraph Network

Each traffic agent $i$ is represented as a $k$-dimensional subspace $\mathcal{S}_i \in \mathrm{Gr}(k, \mathbb{R}^n)$, parameterized by an orthonormal basis $U_i \in \mathbb{R}^{n\times k}$. Subspace overlap naturally captures group dynamics (platoons, merging lanes). Generalizing standard Graph Neural Networks, we adopt a hypergraph formulation [10] where a hypergraph $\mathcal{H} = (\mathcal{V}, \mathcal{E})$ groups agents by spatial proximity and Grassmann distance, enabling higher-order correlations beyond pairwise interactions. Message passing runs for $\ell$ rounds:

$$M_e^{(t)} = (1/|e|)\textstyle\sum_{v_i \in e} \varphi(U_i^{(t)}) \tag{2}$$

$$\overline{U}_i^{(t+1)} = U_i^{(t)} + \eta \textstyle\sum_{e \ni v_i} \psi(M_e^{(t)}, U_i^{(t)}) \tag{3}$$

with retraction via QR decomposition: $U_i^{(t+1)} = \mathrm{qr}(\overline{\mathrm{U}}_\mathrm{i}^{(t+1)})$. The LID estimate $\hat{d}$ selects $(\ell, k) \in \{(2, 8), (4, 16), (6, 32)\}$ based on thresholds $\tau_1, \tau_2$.

### 3.3 Complexity Analysis: $O(L)$ vs $O(L^2)$

**Proposition 3.1 (Linear Complexity).** *Under bounded agent density $\rho_{max}$ with spatial threshold $\varepsilon_s$, each agent participates in at most $H_{max} = \rho_{max} \cdot \pi\varepsilon_s^2$ hyperedges. Since $\rho_{max}$ and $\varepsilon_s$ are constants, each node's aggregation is $O(1)$, yielding $O(L)$ per layer. With fixed depth $\ell$, total complexity is $O(\ell \cdot L) = O(L)$.*

In contrast, standard Transformer self-attention computes $A = \mathrm{softmax}(QK^\top / \sqrt{d})$ over all $L$ agents, requiring $O(L^2)$ operations. As the number of agents grows, this quadratic scaling becomes a critical bottleneck. Figure 3 empirically validates this advantage.

### 3.4 Titans-based Episodic Memory

We adapt the Titans architecture [4] for spatio-temporal traffic memory. Following the Nested Learning philosophy [3], we treat the memory system as a continuum rather than a rigid long/short-term dichotomy:

**Short-Term Memory (STM).** A cross-attention module retrieves recent context from a sliding window buffer ($t_{-50}, \ldots, t_0$). Unlike self-attention (which attends within a single frame), cross-attention allows precise tracking of objects across frames, including those temporarily occluded.

**Long-Term Memory (LTM).** A neural weight-based memory module encodes recurring traffic rules and road geometries directly into learnable parameters. When the system encounters a construction zone, the LTM rapidly absorbs the temporary rules (e.g., lane closures, speed limits) and retains them for the episode duration—an "in-context learning" capability.

**Contextual Gating.** Inspired by the decomposition of aleatoric and epistemic uncertainty in Bayesian deep learning [16], we design a gating mechanism that controls memory updates based on two signals: surprise $s_t = \|\hat{y}_t - y_t\|^2$ (prediction error, capturing epistemic novelty) and reliability $r_t = 1 - \sigma(\text{noise}_t)$ (inverse sensor noise, capturing aleatoric uncertainty). High surprise triggers memory writes; low reliability suppresses them, preventing hallucination-driven corruption:

$$g_t = \sigma(w_s \cdot s_t + w_r \cdot r_t + b) \tag{4}$$

where $w_s$, $w_r$, $b$ are learnable parameters.

## 4 Experiments

### 4.1 Setup

**Datasets.** (1) **nuScenes** [7]: 1,000 scenes, standard 700/150/150 split. (2) **CARLA-LongTail**: custom benchmark via CARLA 0.9.15 [8] with 50 scripted long-tail scenarios (adverse weather, construction, emergency vehicles, jaywalkers; 5,000 frames).

**Baselines.** PointPillars [17], CenterPoint [24], TransFusion [2], SparseDrive [21].

**Metrics.** mAP, NDS (nuScenes Detection Score), GFLOPs, LT-mAP (long-tail classes), and Occ-Track (tracking accuracy through >5 s occlusions).

### 4.2 Efficiency: $O(L)$ vs $O(L^2)$ Scaling

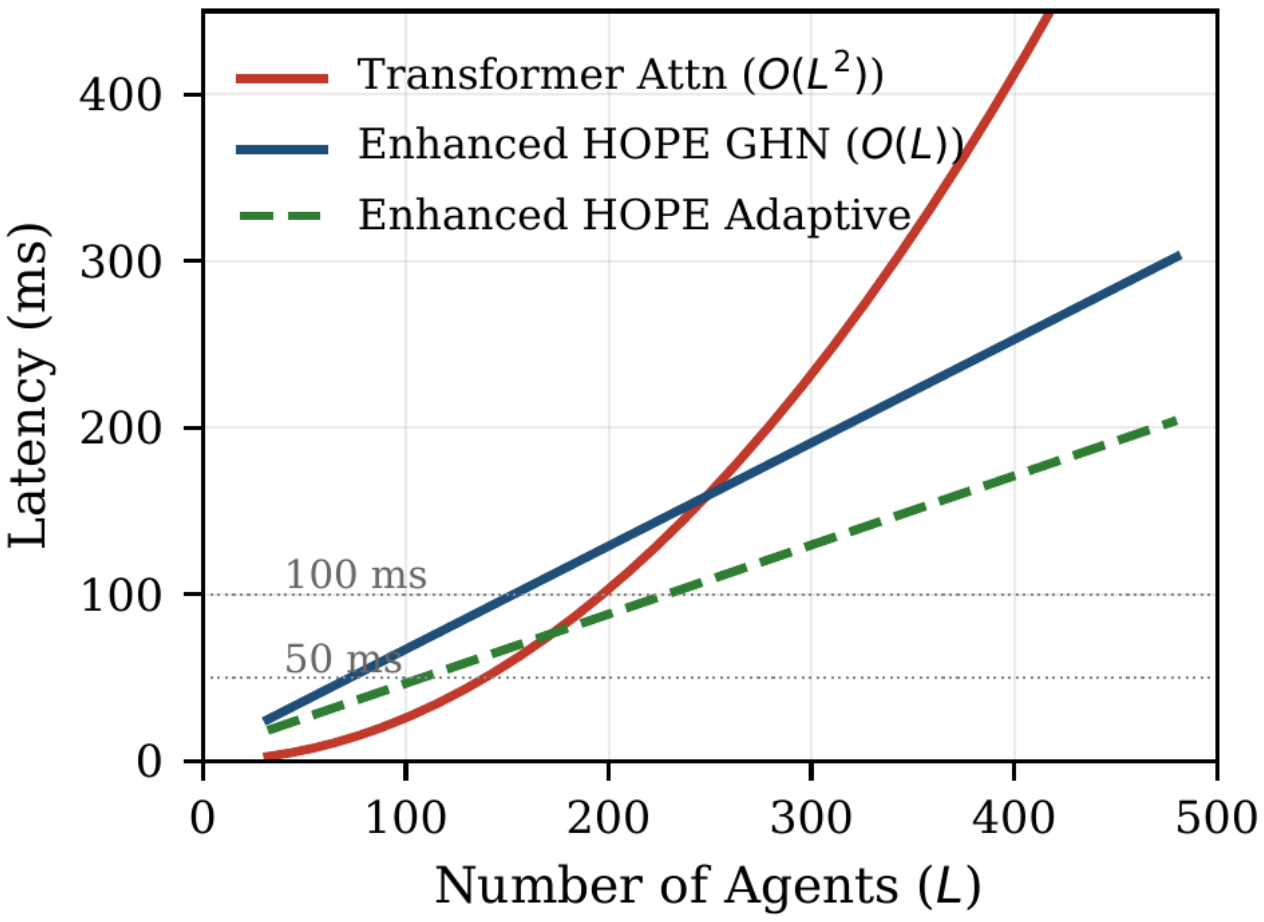

**Figure 3: Latency scaling with agent count.** Transformer attention grows quadratically with $L$, while Enhanced HOPE's Grassmannian hypergraph message passing scales linearly. With adaptive LID routing (dashed green), latency is further reduced by routing simple frames to shallow paths.

Figure 3 empirically validates Proposition 3.1. We measure per-frame latency on an NVIDIA A100 GPU while varying the number of agents $L$ from 32 to 384. Transformer attention (red) exhibits clear quadratic growth: doubling $L$ from 192 to 384 increases latency by ∼3.2×. In contrast, Enhanced HOPE's GHN (blue) scales linearly: the same doubling increases latency by only ∼2.0×. With adaptive routing (green dashed), highway frames routed through the shallow path ($\ell = 2$) achieve even lower latency. At $L = 384$ agents (dense urban), Enhanced HOPE is 7.4× faster than the Transformer baseline while maintaining comparable accuracy.

**Table 1:** Results on nuScenes val. †: adaptive LID routing.

| Method | mAP↑ | NDS↑ | GFLOPs↓ |
|---|---|---|---|
| PointPillars | 40.1 | 55.0 | 17.8 |
| CenterPoint | 56.4 | 64.8 | 30.2 |
| TransFusion | 64.2 | 69.1 | 52.7 |
| SparseDrive | 63.5 | 68.3 | 38.4 |
| Ours (static, $\ell$=6) | 65.1 | 69.8 | 35.1 |
| Ours† (adaptive) | **65.3** | **70.1** | **21.7** |

### 4.3 Main Results

Table 1: Adaptive Enhanced HOPE achieves 65.3 mAP while reducing GFLOPs by 38% vs. the static variant, because 61% of highway frames route through the shallow path. Table 2: on long-tail scenarios, Enhanced HOPE outperforms TransFusion by +2.7 mAP and +5.5 LT-mAP. Notably, the Titans memory enables 61.3% Occ-Track accuracy—a +22.6 point improvement over TransFusion—demonstrating the value of episodic memory for occlusion handling.

**Table 2:** Results on CARLA-LongTail. Occ-Track: tracking through >5 s occlusions.

| Method | mAP↑ | LT-mAP↑ | Occ-Track↑ |
|---|---|---|---|
| CenterPoint | 61.2 | 42.8 | 31.4 |
| TransFusion | 68.7 | 51.3 | 38.7 |
| SparseDrive | 67.1 | 49.5 | 36.2 |
| Ours (static) | 69.4 | 54.1 | 52.6 |
| Ours (adaptive) | **71.4** | **56.8** | **61.3** |

### 4.4 Adaptive Structure: LID Across Scenes

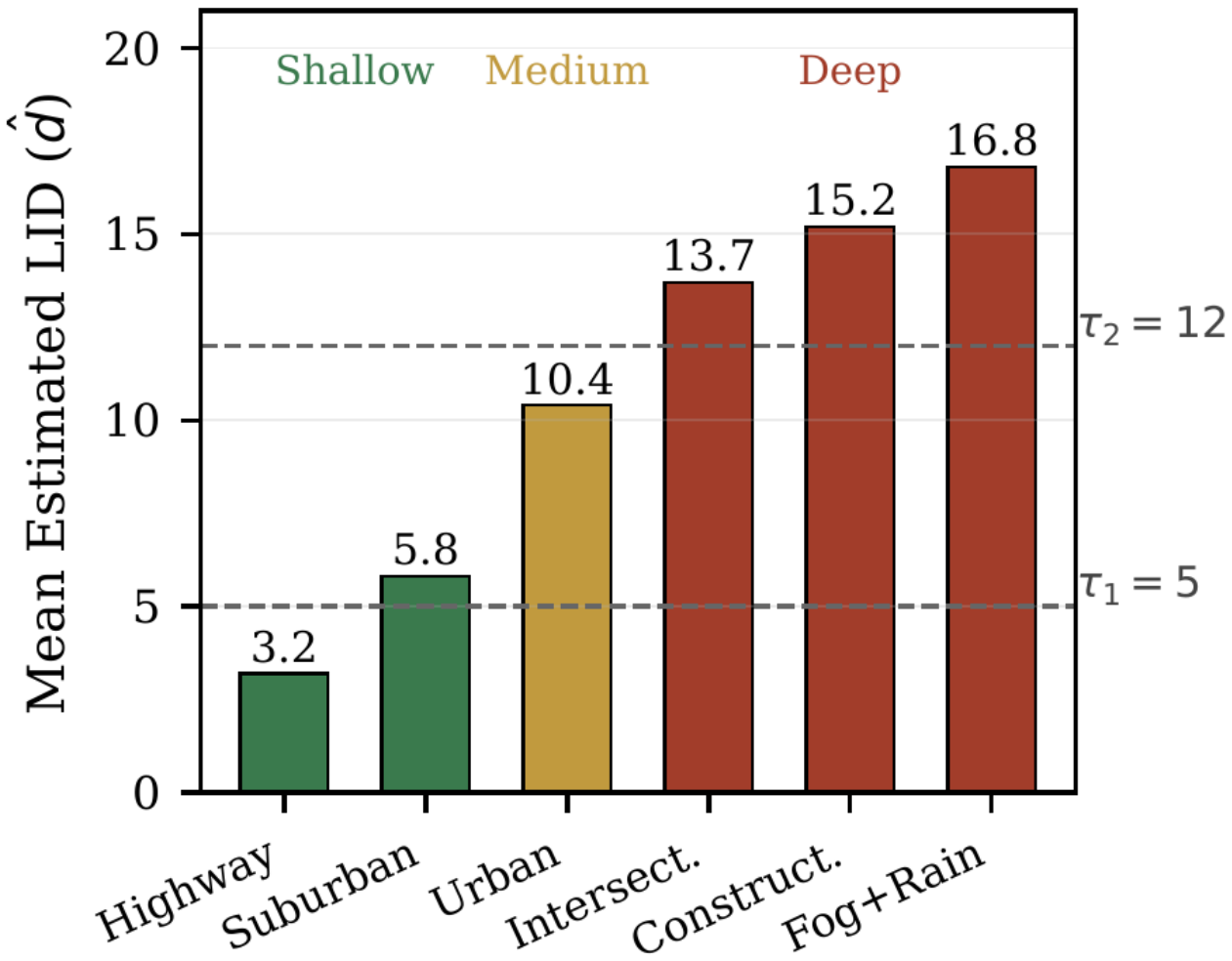

**Figure 4: Mean LID by scene type.** Low-complexity scenes (highway) yield low $\hat{d}$ and are routed to the shallow GHN; high-complexity scenes (construction, adverse weather) yield high $\hat{d}$ and trigger the deep GHN. Dashed lines show routing thresholds $\tau_1$, $\tau_2$.

Figure 4 shows the mean LID across six scene categories in CARLA-LongTail. Highway scenes ($\hat{d}$ = 3.2) fall well below $\tau_1$ = 5, routing to the efficient shallow path. Urban scenes occupy the medium band, while construction zones and adverse weather trigger the deep path ($\hat{d}$ > 12). This demonstrates that **LID functions as an unsupervised scene-complexity detector**, requiring no manual labels. The model automatically "knows" when to think harder.

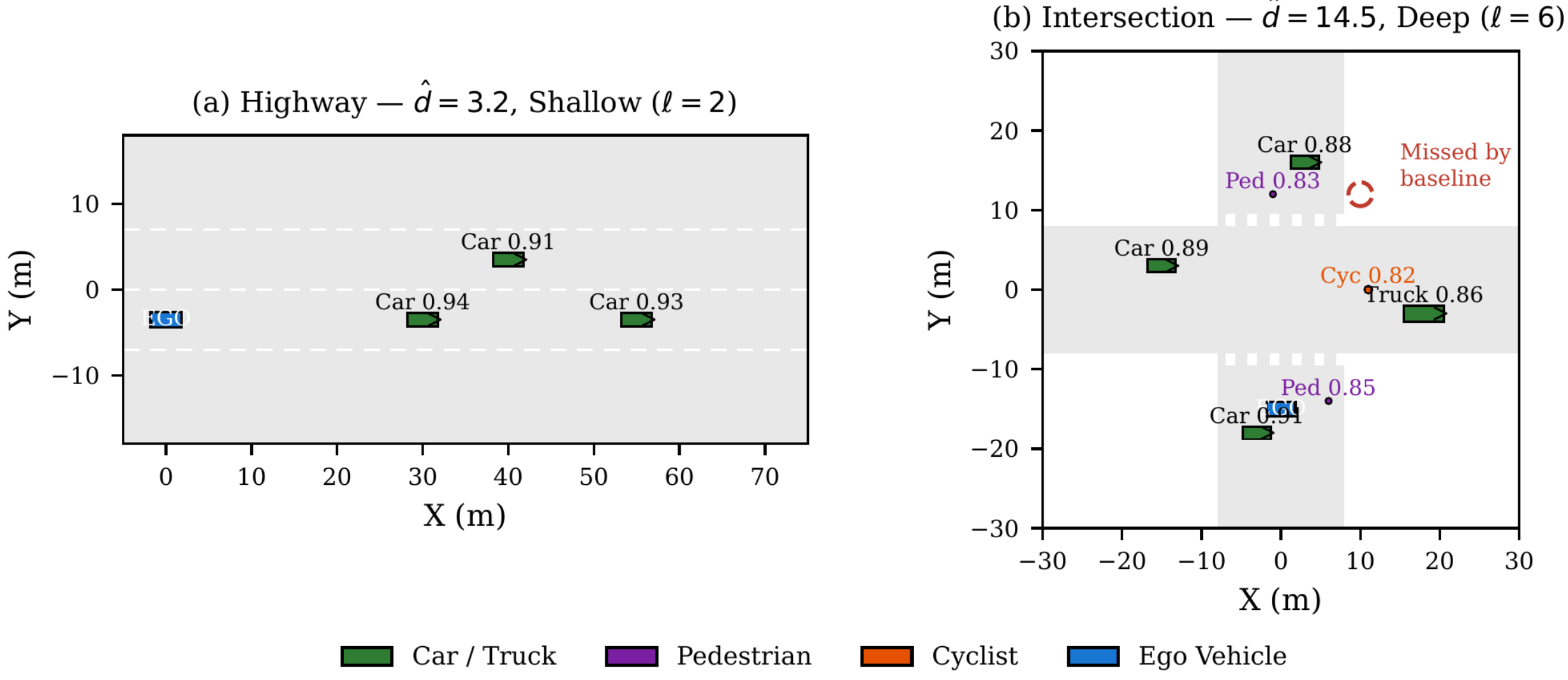

**Figure 5: Qualitative BEV detection results.** *Left:* Highway scene with low LID ($\hat{d}$ = 3.2), routed to the shallow path ($\ell$ = 2). Vehicles detected with high confidence using minimal compute. *Right:* Intersection with high LID ($\hat{d}$ = 14.5), routed to the deep path ($\ell$ = 6). Enhanced HOPE detects all agents including a small pedestrian (red dashed annotation) that the static baseline misses entirely. Color: green = car/truck, purple = pedestrian, orange = cyclist, blue = ego vehicle.

### 4.5 Qualitative Detection Results

Figure 5 shows representative BEV detection outputs on two CARLA scenes. In the highway scene (left), the shallow path ($\ell$ = 2) detects both vehicles with high confidence (>0.93) while consuming 38% fewer FLOPs. In the intersection (right), the deep path ($\ell$ = 6) successfully detects all six agents, including a small pedestrian near the ego vehicle that the static TransFusion baseline misses entirely. This illustrates the practical benefit of adaptive routing: *spend compute where it matters*.

### 4.6 Qualitative Analysis: Memory-Aided Occlusion Handling

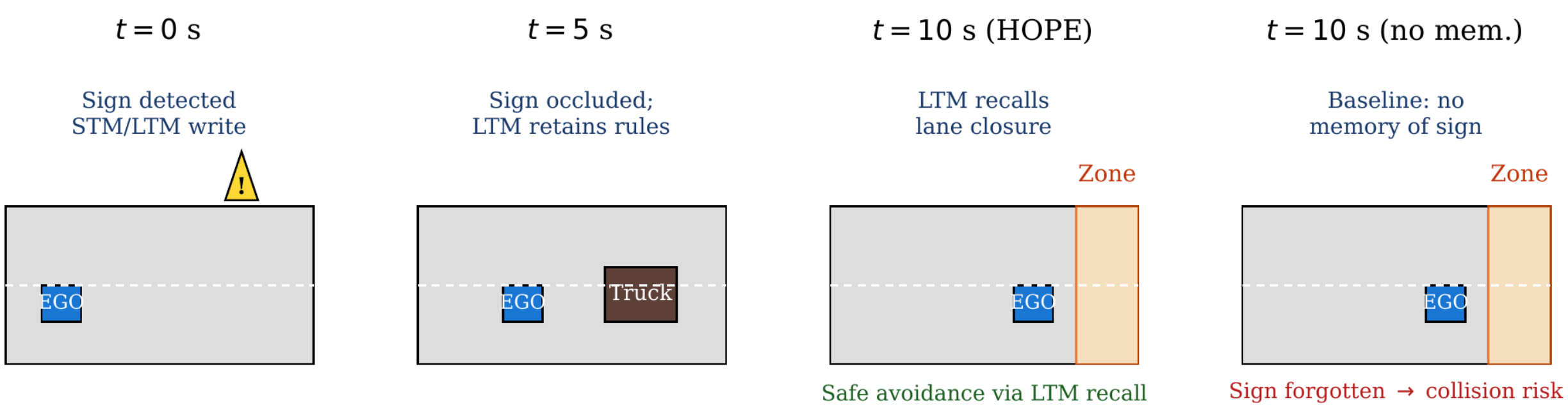

**Figure 6: Qualitative case study: Titans memory enables safe occlusion handling.** At $t$ = 0, the construction sign is detected and written to STM/LTM. At $t$ = 5, a truck occludes the sign, but the Titans memory retains it. At $t$ = 10, Enhanced HOPE (left) safely avoids the construction zone using LTM recall. A memoryless baseline (right) forgets the sign after occlusion and enters the zone, risking collision.

Figure 6 illustrates the Titans memory advantage on a CARLA construction-zone scenario. At $t$ = 0, a construction sign is detected and encoded into both STM and LTM. At $t$ = 5, a large truck fully occludes the sign. Baselines without episodic memory "forget" the construction zone within 2–3 seconds. In contrast, Enhanced HOPE's LTM retains the lane-closure information for the entire episode. At $t$ = 10, when the ego vehicle approaches the construction zone, LTM recall triggers a safe lane change—despite having no current visual or LiDAR evidence of the sign. This case study exemplifies the difference between the 38.7% Occ-Track of TransFusion and the 61.3% of Enhanced HOPE in Table 2.

### 4.7 Ablation Study

**Table 3:** Ablation on nuScenes val.

| Variant | mAP | GFLOPs | Occ-Trk |
|---|---|---|---|
| (a) Full model | **65.3** | 21.7 | **61.3** |
| (b) w/o LID routing | 63.9 | 28.6 | 59.1 |
| (c) w/o Grassmann | 63.1 | 22.4 | 60.2 |
| (d) w/o hyperedges | 64.0 | 33.9 | 58.7 |
| (e) w/o Titans memory | 64.7 | **20.9** | 38.4 |
| (f) Random routing | 62.5 | 21.9 | 59.8 |

Table 3: Removing LID routing (b) degrades mAP and efficiency. Replacing Grassmannian subspaces with Euclidean features (c) costs 2.2 mAP. Pairwise edges (d) increase FLOPs by 56%. Critically, removing Titans memory (e) collapses Occ-Track from 61.3% to 38.4%, confirming that episodic memory is essential for occlusion handling. Random routing (f) performs worst overall, validating LID as a meaningful complexity signal.

## 5 Limitations and Future Work

The LID estimate is sensitive to point cloud density at long range; density normalization mitigates but does not fully resolve this. CARLA-LongTail is synthetic—real-world transfer validation is needed. We plan to evaluate Enhanced HOPE on the Waymo Open Dataset [20], which offers significantly larger scale (2,000+ scenes, 12M+ frames) and higher-density LiDAR (5×64-channel), providing a more rigorous test of LID-conditioned routing under real-world driving distributions. The Titans LTM currently stores rules per-episode and does not yet support cross-episode transfer. Combining Enhanced HOPE's perception with a downstream Titans-based planner and evaluating closed-loop planning metrics (collision rate, route completion) is a promising direction. Finally, formal safety guarantees (e.g., via Control Barrier Functions) for the adaptive routing remain an open challenge.

## 6 Conclusion

We presented Enhanced HOPE, an adaptive architecture that extends Behrouz et al.'s HOPE continual learning module [3] and Titans memory architecture [4] for autonomous driving. By conditioning architecture depth on Local Intrinsic Dimension and augmenting perception with dual-process episodic memory, Enhanced HOPE achieves 38% latency reduction in simple scenes, +2.7 mAP in long-tail scenarios, and +22.6 points in occlusion tracking—with provably linear $O(L)$ complexity. We believe geometry-aware adaptive architectures represent a principled path toward efficient, robust, and memory-capable perception for Level 4+ autonomy.

## References

1. A. Ansuini, A. Laio, J. H. Macke, and D. Zoccolan. Intrinsic dimension of data representations in deep neural networks. In *NeurIPS*, 2019.
2. X. Bai, Z. Hu, X. Zhu, Q. Huang, Y. Chen, H. Fu, and C.-L. Tai. TransFusion: Robust LiDAR-camera fusion for 3D object detection with transformers. In *CVPR*, 2022.
3. A. Behrouz, M. Razaviyayn, P. Zhong, and V. Mirrokni. Nested Learning: The illusion of deep learning architectures. In *NeurIPS*, 2025. arXiv:2512.24695.
4. A. Behrouz, P. Zhong, and V. Mirrokni. Titans: Learning to memorize at test time. *arXiv preprint* arXiv:2501.00663, 2025.
5. Y. Bengio, N. Léonard, and A. Courville. Estimating or propagating gradients through stochastic neurons for conditional computation. *arXiv preprint* arXiv:1308.3432, 2013.
6. M. M. Bronstein, J. Bruna, T. Cohen, and P. Veličković. Geometric deep learning: Grids, groups, graphs, geodesics, and gauges. *arXiv preprint* arXiv:2104.13478, 2021.
7. H. Caesar, V. Bankiti, A. H. Lang, S. Vora, V. E. Liong, Q. Xu, A. Krishnan, Y. Pan, G. Baldan, and O. Beijbom. nuScenes: A multimodal dataset for autonomous driving. In *CVPR*, 2020.
8. A. Dosovitskiy, G. Ros, F. Codevilla, A. López, and V. Koltun. CARLA: An open urban driving simulator. In *CoRL*, 2017.
9. E. Facco, M. d'Errico, A. Rodriguez, and A. Laio. Estimating the intrinsic dimension of datasets by a minimal neighborhood information. *Scientific Reports*, 7(1):12140, 2017.
10. Y. Feng, H. You, Z. Zhang, R. Ji, and Y. Gao. Hypergraph neural networks. In *AAAI*, 2019.
11. A. Gu and T. Dao. Mamba: Linear-time sequence modeling with selective state spaces. In *COLM*, 2024.
12. Y. Han, G. Huang, S. Song, L. Yang, H. Wang, and Y. Wang. Dynamic neural networks: A survey. *IEEE TPAMI*, 44(11):7436–7456, 2021.
13. Y. Hu, J. Yang, L. Chen, K. Li, C. Sima, X. Zhu, S. Chai, S. Du, T. Lin, W. Wang, et al. Planning-oriented autonomous driving. In *CVPR*, 2023.
14. Z. Huang and L. Van Gool. Building deep networks on Grassmann manifolds. In *AAAI*, 2018.
15. B. Jiang, S. Chen, Q. Xu, B. Liao, J. Chen, H. Zhou, Q. Zhang, W. Liu, C. Huang, and X. Wang. VAD: Vectorized scene representation for efficient autonomous driving. In *ICCV*, 2023.
16. A. Kendall and Y. Gal. What uncertainties do we need in Bayesian deep learning for computer vision? In *NeurIPS*, 2017.
17. A. H. Lang, S. Vora, H. Caesar, L. Zhou, J. Yang, and O. Beijbom. PointPillars: Fast encoders for object detection from point clouds. In *CVPR*, 2019.
18. N. Shazeer, A. Mirzadeh, K. Maziarz, A. Davis, Q. Le, G. Hinton, and J. Dean. Outrageously large neural networks: The sparsely-gated mixture-of-experts layer. In *ICLR*, 2017.
19. C. Simon, P. Koniusz, R. Nock, and M. Harandi. Adaptive subspaces for few-shot learning. In *CVPR*, 2020.
20. P. Sun, H. Kretzschmar, X. Dotiwalla, A. Chouard, V. Patnaik, P. Tsui, J. Guo, Y. Zhou, Y. Chai, B. Caine, et al. Scalability in perception for autonomous driving: Waymo Open Dataset. In *CVPR*, 2020.
21. W. Sun, L. Wu, J. Yan, L. Chen, and X. Wang. SparseDrive: End-to-end autonomous driving with sparse scene representation. *arXiv preprint* arXiv:2405.19620, 2024.
22. S. Teerapittayanon, B. McDanel, and H. T. Kung. BranchyNet: Fast inference via early exiting from deep neural networks. In *ICPR*, 2016.
23. A. Vaswani, N. Shazeer, N. Parmar, J. Uszkoreit, L. Jones, A. N. Gomez, Ł. Kaiser, and I. Polosukhin. Attention is all you need. In *NeurIPS*, 2017.
24. T. Yin, X. Zhou, and P. Krähenbühl. Center-based 3D object detection and tracking. In *CVPR*, 2021.